\documentclass[letterpaper]{article} 
\usepackage{aaai2026}  
\usepackage{times}  
\usepackage{helvet}  
\usepackage{courier}  
\usepackage[hyphens]{url}  
\usepackage{graphicx} 
\usepackage{natbib}  
\usepackage{caption} 
\usepackage{algorithm}

\usepackage{newfloat}
\usepackage{listings}
\DeclareCaptionStyle{ruled}{labelfont=normalfont,labelsep=colon,strut=off} 
\floatstyle{ruled}
\newfloat{listing}{tb}{lst}{}
\floatname{listing}{Listing}
\usepackage{multirow}
\usepackage{xcolor, color, colortbl}
\usepackage{algorithm}
\usepackage{algorithmicx}
\definecolor{DPgreen}{rgb}{0.90, 1.0, 0.8}
\definecolor{UPblue}{rgb}{0.80, 1.0, 1.0}
\usepackage{subcaption}
\usepackage{amssymb,amsthm,amsmath,amsfonts}
\newcommand\scalemath[2]{\scalebox{#1}{\mbox{\ensuremath{\displaystyle #2}}}}

\title{Preserving Topological and Geometric Embeddings for Point Cloud Recovery}
\author {
    Kaiyue Zhou\textsuperscript{\rm 1,2},
    Zelong Tan\textsuperscript{\rm 1,2},
    Hongxiao Wang\textsuperscript{\rm 3},
    Ya-Li Li\textsuperscript{\rm 1,2},
    Shengjin Wang\textsuperscript{\rm 1,2}\thanks{Corresponding author: wgsgj@tsinghua.edu.cn}
}
\affiliations {
    \textsuperscript{\rm 1}Department of Electronic Engineering, Tsinghua University\\
    \textsuperscript{\rm 2}Beijing National Research Center for Information Science and Technology (BNRist)\\
    \textsuperscript{\rm 3}Academy for Multidisciplinary Studies, Capital Normal University\\
}

\usepackage{bibentry}

\begin{document}

\maketitle

\begin{abstract}
Recovering point clouds involves the sequential process of sampling and restoration, yet existing methods struggle to effectively leverage both topological and geometric attributes. To address this, we propose an end-to-end architecture named \textbf{TopGeoFormer}, which maintains these critical properties throughout the sampling and restoration phases. 
First, we revisit traditional feature extraction techniques to yield topological embedding using a continuous mapping of relative relationships between neighboring points, and integrate it in both phases for preserving the structure of the original space.
Second, we propose the \textbf{InterTwining Attention} to fully merge topological and geometric embeddings, which queries shape with local awareness in both phases to form a learnable 3D shape context facilitated with point-wise, point-shape-wise, and intra-shape features.
Third, we introduce a full geometry loss and a topological constraint loss to optimize the embeddings in both Euclidean and topological spaces. The geometry loss uses inconsistent matching between coarse-to-fine generations and targets for reconstructing better geometric details, and the constraint loss limits embedding variances for better approximation of the topological space.
In experiments, we comprehensively analyze the circumstances using the conventional and learning-based sampling/upsampling/recovery algorithms. The quantitative and qualitative results demonstrate that our method significantly outperforms existing sampling and recovery methods.
\end{abstract}

\begin{links}
    \link{Code}{https://github.com/ky-zhou/TopGeoFormer}
\end{links}

\section{Introduction}
\label{sec:intro}

Current 3D scanning technologies are essential for capturing detailed representations of objects, environments, or even entire landscapes in the form of dense point clouds. They generally achieve high accuracy and resolution, especially for small objects. Here, we define \textbf{recovery} as the process of both downsampling and subsequent restoration of a point cloud. To integrate data from various remote devices efficiently, point cloud recovery becomes a fundamental methodology in various applications, ranging from remote robot maneuver, augmented reality interaction, to medical surgery interaction. As an emerging field, it is different from point cloud completion, because the focus shifts from dealing with imperfection to efficiently sampling and restoring the point cloud while minimizing information loss. Moreover, point cloud completion requires transmitting the entire input directly to the decoding phase, which inherently conflicts with the purpose defined in recovery. In contrast to compression~\cite{schwarz2018emerging} focusing solely on transmission efficiency, recovery aims to jointly optimize both geometric fidelity and transmission efficiency.

\begin{figure}[t]
  \centering
  \includegraphics[width=.94\linewidth]{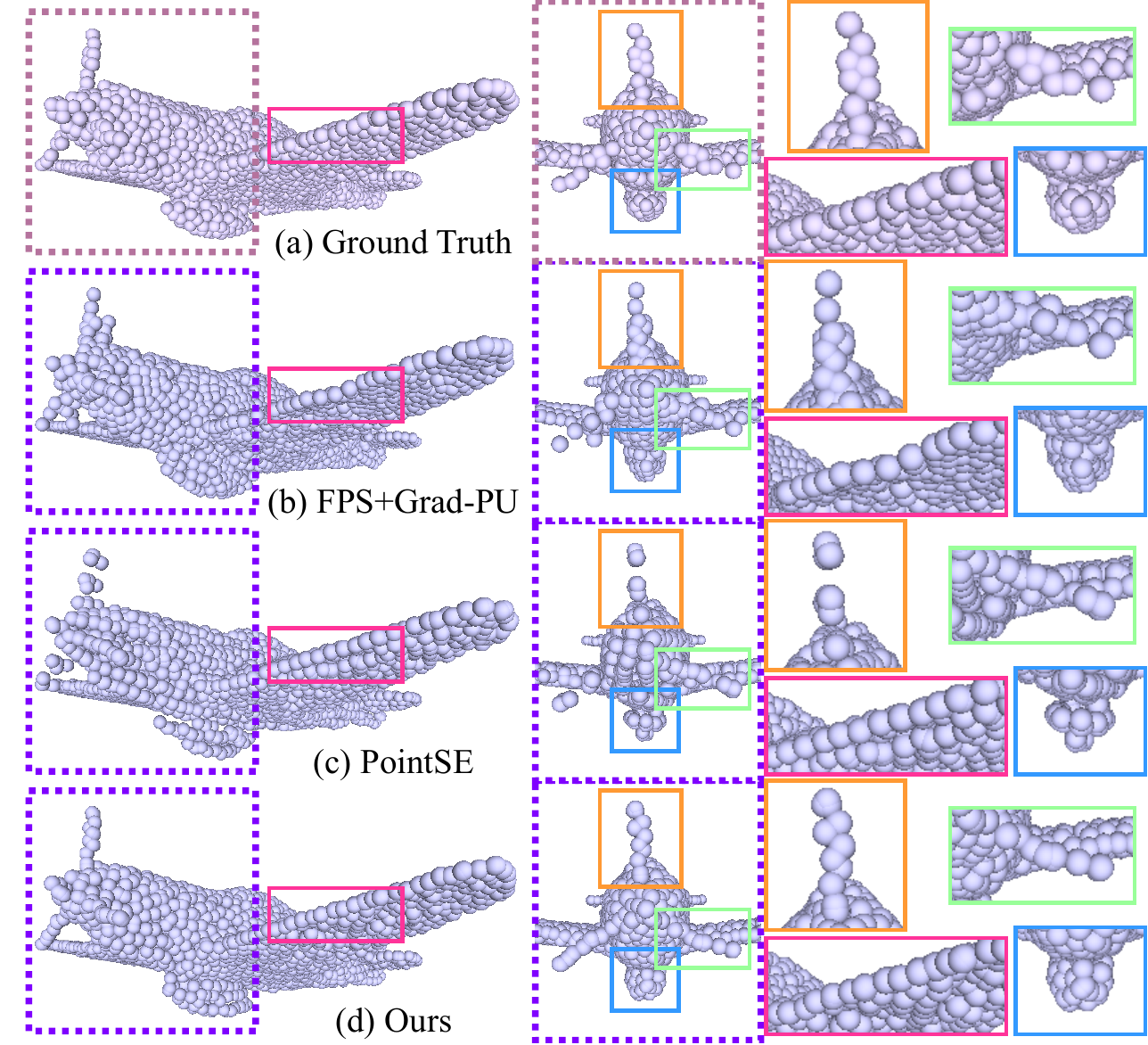}
  \caption{Overview of an aircraft and front views of recovered details. More details such as continuity and curvature (propeller, fuel tank, and wing) are preserved by our method.}
  \label{fig:show}
\end{figure}

The quality of recovered points depends on the procedure of sampling and restoration.
An intuitive paradigm generally utilizes a traditional downsampling algorithm, farthest point sampling (FPS), cascaded by a followed upsampling method for this task. However, such a two-stage pipeline may lead to the loss of fine details due to the nature of FPS, which aims to uniformly sample points on a surface. On the contrary, the sampling process is essential to the overall reconstruction accuracy, since point cloud sampling must maintain the integrity of spatial relationships between points while balancing geometric fidelity with data volume. This is more challenging than image downsampling techniques due to the orderless nature of point clouds. 

To compensate for this tradeoff, existing point cloud recovery methods tend to learn displacements based on FPS-sampled points to capture the geometric pattern.
Point Set Self-Embedding~\cite{li2022point} (PointSE) was first proposed to explore this domain, using an end-to-end paradigm for sampling and restoring objects represented by point clouds. However, it directly formulates the topological features using many k-nearest-neighbors (kNN) operations without fully leveraging that rich information with the shape, resulting extremely more expensive computation and less effective results.
Meanwhile, the Chamfer distance (CD) loss is applied on recovered point clouds with their ground truth counterparts in a consistent manner, producing predictions with fewer local geometric details. 

In this paper, we focus on the extremely under-explored research field, i.e., point cloud recovery, by proposing \textit{TopGeoFormer} for preserving both topological and geometric attributes. First, we formulate a Down-Preservation module generating a 3D shape context to capture intra-shape features. Instead of directly mapping the 1D global code to the Euclidean space, we devise the InterTwining Attention (ITA) that queries a general learnable 2D shape code with point-wise features guided by topological embeddings to contextually encode local geometric patterns. Second, we propose an Up-Preservation phase, which integrates ITA, Up-Preserving Attention (UPA), and residual multi-layer perceptrons (MLPs), aiming to restore dense points. Third, to learn explicit local details, we propose to optimize the network based on the inconsistent mappings between the restored point clouds at various resolutions and the ground truth. Comparing to consistently matching the ground truth resolution with generations, this strategy effectively refines the local details of the coarse point clouds produced by intermediate layers. Lastly, we propose a topological constraint loss to guide the optimization process, which limits embedding variance in the latent space and slightly enhances recovery performance. By such designs, we achieve superior performance compared to state-of-the-arts by a large margin in terms of both sampling and recovery tasks. Qualitative results indicate that sampled and restored objects exhibit significantly more original and realistic local geometry. We summarize our contributions as follows:

\begin{itemize}
    \item We revisit traditional feature extraction to fully explore the utility of topological embedding, which is subtraction relations between a local centroid and its neighbors. Such embedding is essential to guide intra-shape features for precise sampling and restoration. 
    \item To merge topological and geometric embeddings, we propose intertwining attention by interacting local and point-wise features with the rough global shape code, which can be transformed to intra-shape features to further enrich the current shape information for closely resembling the original structure.
    \item Our losses are specifically designed for the proposed paradigm with respect to both Euclidean and topological spaces for preservation of fine-details. The full geometry loss always utilizes ground truth to guide coarse-to-fine predictions, and the topological constraint loss ensures the encoded embeddings to remain coherent with their original local structures. 
\end{itemize}

\section{Related Work}\label{sec:relate}

\noindent\textbf{Point Cloud Sampling. }
Point cloud sampling (downsampling) methods~\cite{li2018so,dovrat2019learning,lang2020samplenet,wu2023attention} are generally integrated with understanding and reconstruction downstreams. Our method can be categorized as reconstruction-oriented sampling that preserves shape details without any semantic priors~\cite{liu2024lta}.

\noindent\textbf{Point Cloud Upsampling. }
Point cloud upsampling falls into three main types: shuffle-based~\cite{yu2018pu,qian2021pu,luo2021pu,ye2021meta,qiu2022pu}, tile-based~\cite{yifan2019patch,li2019pu,li2021point,rong2024repkpu}, and others based on implicit functions~\cite{ma2021neural,feng2022neural,zhao2022self,he2023grad}, diffusion models~\cite{qu2024conditional}, or other modalities~\cite{qian2020pugeo,zhao2021sspu,liu2024spu,yang2024tulip}. This task generally restores uniform points and fills small holes, which differentiates from recovery that focuses more on restoring details.

\begin{figure*}[h]
\begin{center}
    \includegraphics[width=1.0\linewidth]{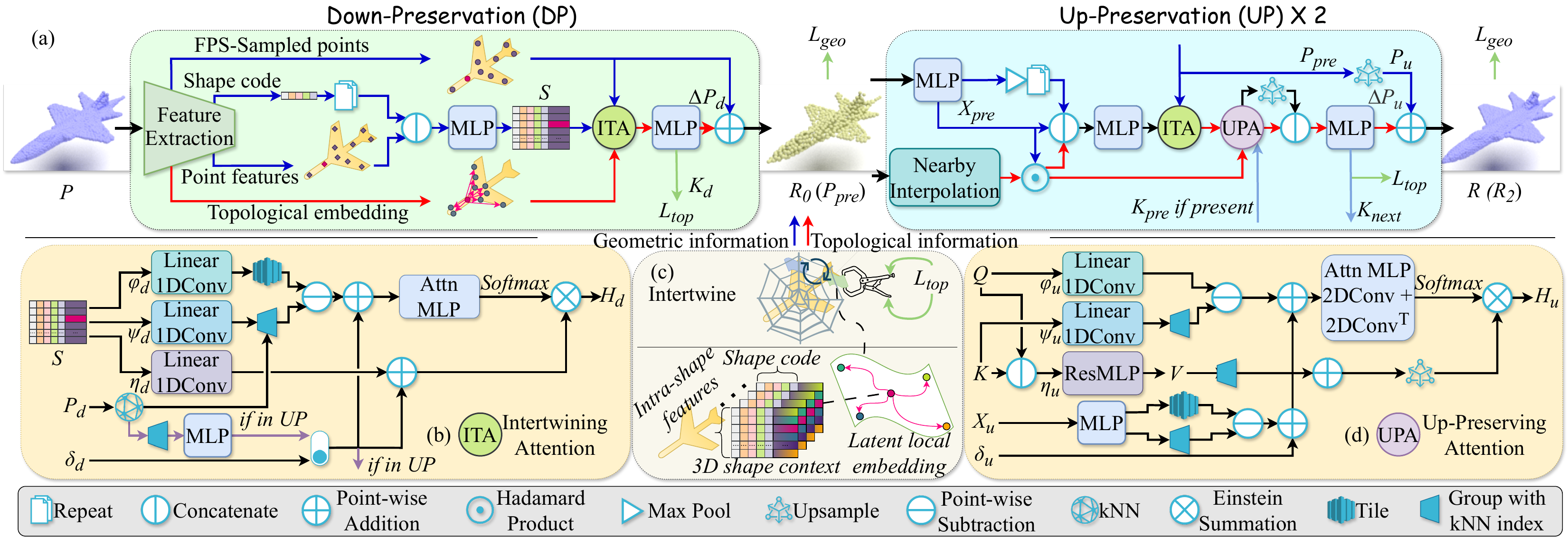}
\end{center}
   \caption{Architecture of our \textit{TopGeoFormer} (a) consists of a Down-Preservation (DP) phase and two cascaded Up-Preservation (UP) phases, to preserve both topological and geometric attributes. (b) The learnable 2D shape code $S$ is (c) intertwined with latent embeddings to reflect tangent-space structures. (d) Assisted with ITA, UPA preserves similar features for restoration.}
\label{fig:arch}
\end{figure*}

\noindent\textbf{Point Cloud Completion. }
Point cloud completion is analogous to recovery by merging restoration and upsampling into an individual network. However, serving as a rough guidance, the partial input is inappropriate for recovery where the input that is the entire point cloud should remain unknown during recovery. 
Without other priors rather than coordinates, PCN~\cite{yuan2018pcn} was first proposed to enlighten many classic methods~\cite{tchapmi2019topnet,wen2021pmp,xiang2021snowflakenet,cui2023p2c}. Recent methods are developed using transformers~\cite{yu2021pointr,zhou2022seedformer,li2023proxyformer,chen2023anchorformer,zhu2023svdformer} or semantic information~\cite{xu2023casfusionnet,xiang2023retro,xia2023scpnet,xu2023cp3,zhou2024cascaded}.

\noindent\textbf{Point Cloud Recovery. }
Pioneering the realm of recovery, PointSE presents an end-to-end network to sample and restore point clouds simultaneously~\cite{li2022point}. As the only method in the community, it proposes that such a paradigm is well suited for long-range communication scenarios, where processing or transmitting the entire point cloud can be impractical due to data volume constraints. 

\section{Methodology of TopGeoFormer}
In this work, we address the task of point cloud recovery while preserving the target's topological and geometric properties. As the overview shown in Fig.~\ref{fig:arch}, our network has an end-to-end architecture, comprising a down-preservation (DP) phase and two up-preservation (UP) phases. Given a point set $P = \{ p_i | i=1,...,N\} \in \mathbb{R}^{N \times 3}$, we aim to learn a mapping function $f$ that transforms $p \in P$ continuously into a sampled point cloud $R_0$ representing the coarse shape that closely resembles the original shape $P$, and then into a restored point cloud $R$. Such transformation process can preserve the topological and geometric properties if the mapping function $f$ is continuous and the local Euclidean neighborhoods exhibit sufficient overlap across submanifolds. 
Therefore, we aim to improve recovery capacity in terms of shape details based on \textbf{topological} and \textbf{geometric} attributes. 
We name our approach as the Topology-Geometry Preserving Transformer (\textit{TopGeoFormer}), even though it is inherently a local method~\cite{ma2012manifold}.

\subsection{Down-Preservation}\label{sec:dp} 

Accurate downsampling plays a crucial role in precise object recovery, as the shape code $s$ (typically encoded by PointNet-like~\cite{qi2017pointnet} structures~\cite{qi2017pointnet2, wang2019dynamic}) provides essential guidance for determining a general destination where a seed should move to. 
Generally, $s$ encodes partial topological and geometric information of the manifold $M$. However, directly mapping a shape code $s$ to formulate seeds may introduce additional noise~\cite{xiang2021snowflakenet}. Moreover, simply concatenating $s$ with the non-differentially sampled points still limits the comprehensive awareness on local details, since gradients cannot flow back fluently to update local features.

\noindent\textbf{Extracting Topological and Geometric Embeddings. }
To preserve more local information, instead of only mapping the shape code to a coarse point cloud, we revisit PointNet++~\cite{qi2017pointnet2} and DGCNN~\cite{wang2019dynamic} to additionally reuse the topological information formulated during shape code extraction. Fig.~\ref{fig:arch}(a) shows the structure of our DP phase. Given $P$, we extract the shape code $s \in  \mathbb{R}^{1 \times C_3}$ through a feature extraction module consisting of three set abstraction (SA) layers. Each SA layer is composed of FPS and local grouping based on k-nearest-neighbors (kNN), generating FPS-sampled points $P_d \in  \mathbb{R}^{N_d \times 3}$ as the geometric embedding and their corresponding features $X_d \in  \mathbb{R}^{N_d \times C_1}$ as
\begin{equation}\label{eq:xd}
\scalemath{0.91}{  
    X_d = \mathit{MLP}(\mathit{Cat}(P_d,\delta_d)), \delta_d =\mathit{MLP}(\mathcal{D}),
}
\end{equation}
where $\mathcal{N}_P (p_i)$ denotes the $k$ neighbors of $p_i \in P$, the subtraction operation $\mathcal{D}_i = \{ p_i - p_j| p_j \in \mathcal{N}_P (p_i) \}$ in Euclidean space yields a vector that represents both the magnitude and direction between points, forming the foundation for topological embedding. The subscript $d$ in Eq.~\ref{eq:xd} denotes the first SA layer, representing the set that non-differentially sampled points belong to. 
We map $\mathcal{D} = \{ \mathcal{D}_i \}_{i=1}^{N_d} \in \mathbb{R}^{N_d \times k \times 3}$ to a higher-dimensional latent space $\mathbb{R}^{N_d \times k \times m}$ (where $m > 3$) using an MLP as the continuous map. The MLP can capture complex and fine-grained topological structures, such that $X_d$ preserves the key topological embeddings (e.g., connectivity) of the original space.
Finally, a max-pooling operator aggregates features in $X_d$ along the neighborhood dimension, achieving topological invariant while capturing prominent topological features and resisting to small local perturbations.

During feature extraction, we set the downsampling ratios for the SA layers to 4, 16, and $N$, respectively, and retain $P_d$ and $X_d$ from the first SA for subsequent steps. Preserving local nearest neighbors ensures that the underlying manifold of a point cloud is respected when encoding the data into a different (latent) space during dimensionality reduction. Therefore, we concatenate $X_d$ with $s$ and feed it into a residual MLP ($\mathit{ResMLP}$) to formulate a 2D point-shape fusing code $S \in \mathbb{R}^{N_d \times C_1}$ that comprehensively mixes the local awareness of each point with the overall shape:
\begin{equation}\label{eq:shapecode}
\scalemath{0.91}{  
    S = \mathit{ResMLP}(\mathit{Cat}(X_d,\mathit{Repeat}(s)),
}
\end{equation}
where $s$ is duplicated to fit the size of $X_d$. By merging shape information, $S$ also makes the network permutation-invariant, an essential property for point cloud recovery, given that points may be transmitted asynchronously from different remote sources. Moreover, $S$ serves as a compact latent encoding of the manifold $M$ in the DP phase, which is designed to preserve the topological properties (such as continuity, connectedness) of the original point set $P$. Thus, $S$ encapsulates the structure of the manifold $M$ onto which the point set $P$ or its subsets $P_d$ are mapped. Essentially, $S$ encodes both the geometric and topological structure of $M$, allowing the reconstruction of the shape in Euclidean space via tangent space approximations that preserve the local structure of the manifold. 

\noindent\textbf{Intertwining Attention. }
In order to establish a point-wise, point-shape, and intra-shape context in the manifold, we propose to feed $S$ into a topology-geometry intertwining attention (ITA) module for mapping the latent embeddings guided by tangent space approximations back to the Euclidean space. This process is illustrated in Fig.~\ref{fig:arch}(b). In ITA, $S$ is first transformed by three linear MLPs, denoted as $\varphi_d$, $\psi_d$, and $\eta_d$, respectively. Simultaneously, the topological embedding $\delta_d$ is used to guide the 2D shape code $S$ that describes a general shape (such as a plane sketch), formulating a 3D shape context that encodes local patterns intertwined with a group of 2D shape codes.
The benefit of this operation is twofold: (i) it reduces redundant neighbor computations; (ii) it preserves exhaustive local pattern with respect to $P$. 
Recall that $X_d$ merely represents a rough measure of connectivity due to dimensionality reduction. Here, we aim to learn a more comprehensive manifold by interacting all intra-shape features (representing a same shape) with local embeddings $\delta_d$ to enrich the shape information. The attention vector $H_d$ is formulated as:
\begin{equation}\label{eq:catt}
\scalemath{0.91}{  
\begin{split}
    & h_i =  \sum_{p_j \in \mathcal{N} (p_i)} \rho \bigl( \gamma (\varphi_d (S) - \psi_d (S) + \delta_d) \bigl)
    \otimes \bigl( \eta_d (S) + \delta_d \bigl),\\
    & H_d = \{h_i\}_{i=1}^{N_{d}} \in \mathbb{R}^{N_{d} \times C},
\end{split}
}
\end{equation}
where $N_d$ is the number of downsampled points, $\rho$ is the $\mathit{Softmax}$ activation function, $\gamma$ is an MLP layer with two linear layers and a $\mathit{ReLu}$ nonlinear activation function, $C$ is the dimension of feature, and all transformations of $S$ are broadcasted or grouped to match the dimension of $\delta_d \in \mathbb{R}^{N_{d} \times k \times C}$.
Finally, we feed these intertwining attention vectors $H_d$ into another $\mathit{ResMLP}$ to regress the displacement vectors $\Delta P_d \in  \mathbb{R}^{N_d \times 3}$:
\begin{equation}\label{eq:map}
\scalemath{0.91}{  
    \Delta P_d = \mathit{tanh}(K_d), K_d=\mathit{ResMLP}(H_d),
}
\end{equation}
which are added back to the non-differentially sampled points $P_d$, eventually endowing differentiability to these points as $R_0 \in  \mathbb{R}^{N_d \times 3}$:
\begin{equation}\label{eq:add}
\scalemath{0.91}{  
    R_0 = P_d + \Delta P_d. 
}
\end{equation}

Our sampling (DP) is designed to locally preserve both topological and geometric embeddings. Since the features of all (sampled) points are differentiable, the encoded features comprehensively reflect the topological structure of a point cloud throughout sampling and restoration phases.

\subsection{Up-Preservation}\label{sec:up}

To generate geometric details while precisely preserving topological properties, we propose to progressively restore the point cloud in a coarse-to-fine manner. We devise this preservation procedure to be two cascaded UP phases. For simplicity, we primarily describe an individual UP in Fig.~\ref{fig:arch} in this section. We denote the input to the UP phase as $P_{pre}$ for general expression, where $P_{pre} = R_0$ in the first UP phase. Inspired by decoders using transposed convolution~\cite{xiang2021snowflakenet, zhou2022seedformer}, we adopt their self-attention layer to predict displacements for upsampled points. Furthermore, we update the upsampling features by interpolating the generated points, and cooperate the proposed ITA as a plug-in component with the Up-Preserving Attention (UPA) in each UP module.

Interpolating features between sparsely distributed seeds and generated points~\cite{zhou2022seedformer} is impractical, since seed features are prohibited from transmitting to the restoration stage. Instead, we dynamically update the points to represent the trend for $p_i \in P_{pre}$ to be upsampled in a small local area. We thus calculate the weights to adjust point-wise features as the geometric embedding $X_{pre}$, which formulates the upsampling feature with topological embedding as:
\begin{equation}\label{eq:interpo}
\scalemath{0.91}{  
X_u = \frac{\boldsymbol{1}^\intercal (\mathcal{D}^* \odot X_{pre})}{\boldsymbol{1}^\intercal \mathcal{D}^*} ,\\
\boldsymbol{1}^\intercal \mathcal{D}^* = \sum_{j \in \mathcal{N}_{P_{pre}}(p_i)} \frac  {1} {p_i - p_j},
}
\end{equation}
where $|\mathcal{N}_{P_{pre}}(p_i)|=3$ and $p_i \in P_{pre}$. Parallelly, $P_{pre}$ is encoded by a point-wise MLP to form $X_{pre}$. Both embeddings are then used to enrich the upsampling information in the cascaded intertwining and up-preserving attentions.

We first use ITA to enhance the query $Q=\mathit{ResMLP}\bigl(\mathit{Cat}(X_u, \mathit{Max}(X_{pre}), X_{pre}) \bigl)$ for the following UPA, and $K=Q$ if not present in the current phase. We hereby substitute $S$ with $\mathit{Max}(X_{pre})$ as the updated 2D shape code. $\delta_u$ and its corresponding kNN indices are calculated, since any kNN cache or local pattern is prohibited from transferring between sampling and restoration phases in the recovery task. Still, we can reuse such kNN cache in the subsequent UPA to avoid duplicated computations. 

\noindent\textbf{Up-Preserving Attention. }
We leverage UPA as illustrated in Fig.~\ref{fig:arch}(d) to learn the attention vector for upsampling while preserving the shape information. Let $\varphi_u$ and $\psi_u$ denote linear 1D convolution ($\mathit{1DConv}$) layers for transforming $Q$ and $K$, and $\eta_u$ a $\mathit{ResMLP}$ for concatenating $Q$ and $K$ to formulate the value vector $V$. Without loss of generality, notations such as $Q$, $K$, $V$, etc., omit the index of the current UP phase. Meanwhile, we use an MLP to map $X_u$ to the same dimension of $V$. With the additional upsampling feature $X_u$, we then use an MLP $\gamma_u$ consisting of a linear 2D layer, a $\mathit{ReLu}$ nonlinear layer, and a 2D transposed convolution layer ($\mathit{2DConv}^\intercal$) to merge topological, upsampling, and intra-shape features, formulating the attention vector $H_u$ in each UP. Akin to Eq.~\ref{eq:map} and \ref{eq:add}, the displacements are predicted using $H_u$. In a coarse-to-fine manner, the obtained $R_l$ can be further used as $P_{pre}$ in the next UP phase. Finally, the restored point cloud $R$ is obtained after two UP phases.

\subsection{Optimization}\label{sec:opt} 

\begin{figure*}[ht]
  \centering
  \includegraphics[width=.96\linewidth]{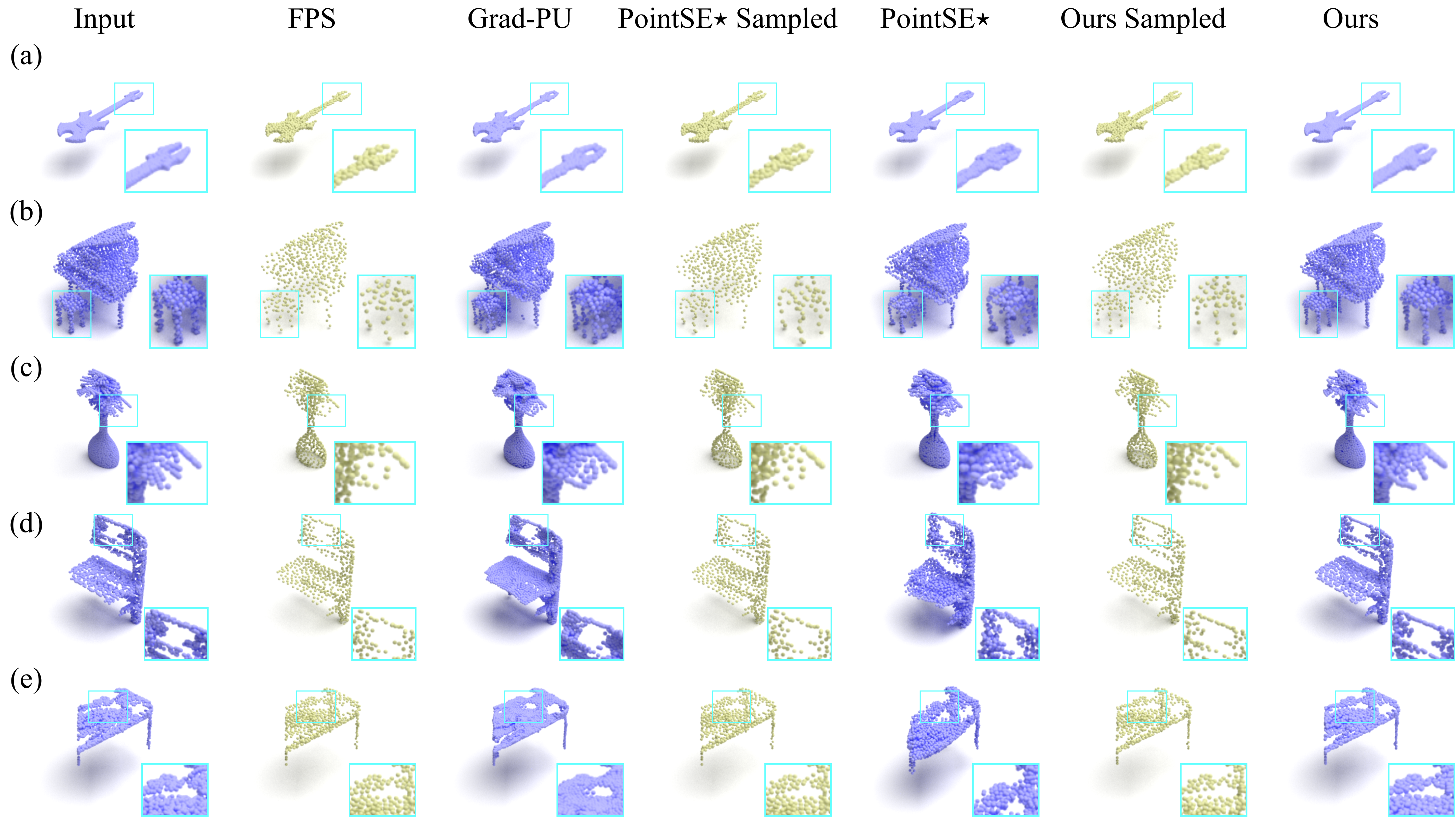}
  \caption{Qualitative visualizations. (a-c) guitar, piano, and plant in the random set. (d-e) chair and table in the partial set. Both our sampling (DP) and recovery methods preserve plausible details and overall shape.}
  \label{fig:visual}
\end{figure*}

Conventional coarse-to-fine methods for point cloud reconstruction~\cite{xiang2021snowflakenet, zhou2022seedformer, zhu2023svdformer} typically leverage progressive CD losses for optimization. 
However, the target point cloud is matched to the same size as the predicted point cloud using FPS, leading to less information preserved during the coarse predictions. To address this, we propose to optimize the coordinates at various resolutions using the original-resolution ground truth as the target. Hence, our full reconstruction loss is formulated as:
\begin{equation}\label{eq:rec}
\scalemath{0.91}{  
\mathcal{L}_\mathit{geo} = \sum_{l=0}^{2} \mathcal{L}_\mathit{CD}(R_l, P),
}
\end{equation}
where $R_1$ is generated by the first UP phase, $R_2 = R$, and $\mathcal{L}_\mathit{CD}$ represents the Chamfer distance (CD), formulated as:
\begin{equation}\label{eq:rec}
\scalemath{0.91}{  
\mathcal{L}_\mathit{CD} = \frac{1}{|P|} \sum_{x \in P} \underset{y \in  R}{\min} {{||x-y||}_2} + \frac{1}{|R|} \sum_{y \in R} \underset{x \in  P}{\min} {{||y-x||}_2},
}
\end{equation}
where ${||\cdot||}_2$ denotes the Euclidean L2-norm of CD and $|\cdot|$ is the magnitude of a matrix. 

According to manifold learning~\cite{belkin2003laplacian}, maintaining small variance in the encoded embeddings implies that the local linear approximations of the manifold (i.e., the tangent spaces) are well-preserved during encoding. This means that local neighborhoods in the original space, which lie on the manifold, remain coherent in the encoded space~\cite{ma2012manifold}.
Therefore, we propose to constrain the encoded embeddings to have small variance as illustrated in Fig~\ref{fig:arch}(c), such that embeddings that differ substantially are limited to overwhelm others. Formally, we denote the topological embedding constraint loss as follows:
\begin{equation}\label{eq:top}
\scalemath{0.91}{  
    \mathcal{L}_\mathit{top} = \sum_{l=0}^{2} \sum_{i=1}^{N_l} \mathit{mean}(K_l^2),
}
\end{equation}
where $N_0=|R_0|=\frac{N}{4}, N_1=|R_1|= \frac{N}{2}, N_2=|R_2|= N, K_0=K_d$. 
Eq.~\ref{eq:top} encourages the topological embeddings to preserve local information mapped into the manifold.

Finally, the overall loss function can be formulated as:
\begin{equation}\label{eq:overall}
\scalemath{0.91}{    
\mathcal{L} = \lambda \mathcal{L}_\mathit{geo} + \mathcal{L}_\mathit{top},
}
\end{equation}
where $\lambda$ is empirically set to 1000. Eq.~\ref{eq:overall} corresponds to optimizations on both geometric and topological levels, redistributing both sampled and recovered points with respect to preserving local shape details.

\section{Experiment}\label{sec:exp}

Following point cloud reconstruction works~\cite{li2022point, li2019pu, qian2021pu, li2021point, he2023grad}, we report metrics: Chamfer distance (CD), Hausdorff distance (HD) using PyTorch3D~\cite{ravi2020pytorch3d}, and Earth mover's distance (EMD) using \cite{liu2020morphing}.
For quantitative experiments, we use: ModelNet40~\cite{wu20153d}, and ScanObjectNN~\cite{uy2019revisiting} to generate three sets: uniform set, random set, and partial set.
For generalization studies and qualitative experiments, we use the collected 10 scenes (KITTI-10) from KITTI~\cite{geiger2013ijrr} and 30 scenes (ScanNet-30) from ScanNet~\cite{dai2017scannet}.
Our network is trained on an Nvidia RTX3090 GPU, with a batch size of 32 for 120 epochs. We use the Adam optimization with initial learning rate 0.005, which is decayed by 0.5 every 30 epochs. During training, point sets are augmented with random mirroring, scaling, and rotation, respectively. Unless otherwise stated, all experiments are conducted with our unified \textit{PyTorch} implementations for fair comparison.
\begin{table}[t]
\begin{center}
\scalebox{.70}{
\begin{tabular}{lccccccccc}
\hline
\multirow{2}{*}{Method} & \multicolumn{3}{c}{Uniform} & \multicolumn{3}{c}{Random} & \multicolumn{3}{c}{Partial} \\
 & CD & HD & EMD & CD & HD & EMD & CD & HD & EMD \\
\hline
PU-GCN  & 1.13 & 1.11 & 6.57 & 1.30 & 1.19 & 7.48 & 0.72 & 0.83 & 6.82 \\
Dis-PU & 0.99 & 1.29 & 6.24 & 1.14 & 1.41 & 7.32 & 0.63 & 0.99 & 6.41 \\
PointSE & 0.76 & 0.74 & 4.51 & 0.90 & 0.85 & 6.45 & 0.47 & 0.49 & 4.63 \\
\hline
RepKPU & 1.00 & 0.36 & 3.09 & 0.99 & 0.35 & 3.07 & 0.51 & 0.24 & 2.63 \\
SampleNet & 1.25 & 1.01 & 7.41 & 1.54 & 1.72 & 9.13 & 0.86 & 1.39 & 10.8 \\
APES & 0.83 & 0.88 & 3.36 & 0.85 & 0.95 & 3.33 & 0.46 & 1.22 & 2.92 \\
PointSE$\star$ & 0.85 & 0.35 & 2.93 & 0.86 & 0.34 & 2.97 & 0.38 & \textbf{0.15} & 2.57 \\
Ours & \textbf{0.63} & \textbf{0.29} & \textbf{2.82} & \textbf{0.62} & \textbf{0.29} & \textbf{2.82} & \textbf{0.26} & 0.18 & \textbf{2.39} \\
\hline
\end{tabular}
}
\end{center}
\caption{Quantitative comparison on recovery in terms of CD ($10^3$), HD ($10^2$), and EMD ($10^2$), between our methods and state-of-the-art recovery and upsampling methods.}\label{tab:recover}
\end{table}

\subsection{Point Cloud Recovery}\label{sec:recovery}


\noindent \textbf{Object-Level. }
Table~\ref{tab:recover} shows the comparisons between our method and other recovery methods, where results in the first 3 rows are directly reported from PointSE (separated by a horizontal line). As the only work in point cloud recovery, we re-implement PointSE in \textit{PyTorch} (denoted as $\star$). For \textit{PyTorch}-based evaluations, we compute the metrics for RepKPU~\cite{rong2024repkpu}, SampleNet~\cite{lang2020samplenet}, APES~\cite{wu2023attention}, PointSE$\star$, and our method. It is worth noting that directly adapting completion methods~\cite{xiang2021snowflakenet, zhou2022seedformer,zhu2023svdformer} into the recovery task generally performs worse than baselines in Table~\ref{tab:recover}, since completion emphasizes more on the decoder with additional guidance from input. Compared to the second best, our method reduces CD on the three sets by 17\%, 27\%, and 32\%, respectively, while also achieving superior performance in terms of almost all metrics. The upsampling methods, PU-GCN, Dis-PU, and RepKPU, use FPS-sampled points as input. 
The result highlights the importance of the sampling ability for shape recovery. Moreover, sampling methods, i.e., SampleNet and APES, generally perform worse than PointSE and ours, even though they are cascaded with the same decoder~\cite{xiang2021snowflakenet} for adjusting from understanding-oriented into a reconstruction-oriented paradigm. Note that, the decoder from SnowflakeNet requires a global shape code for rough seed generation, while ours does not need to transmit any high-dimensional code into the decoding phase. This further strengthens the transmission's efficiency from remote ends.

Meanwhile, we visualize the sampling and corresponding recovery results in Fig.~\ref{fig:visual}. Both our sampled and recovered points better preserve the original details and overall distribution. For instance, in (a-c), our method preserves the guitar's headstock, the chair beside piano, and the leaves of the plant, respectively; and in (d-e), our method maintains either more continuous edges or more accurate holes. 

\begin{figure}[t]
  \centering
  \includegraphics[width=1.0\linewidth]{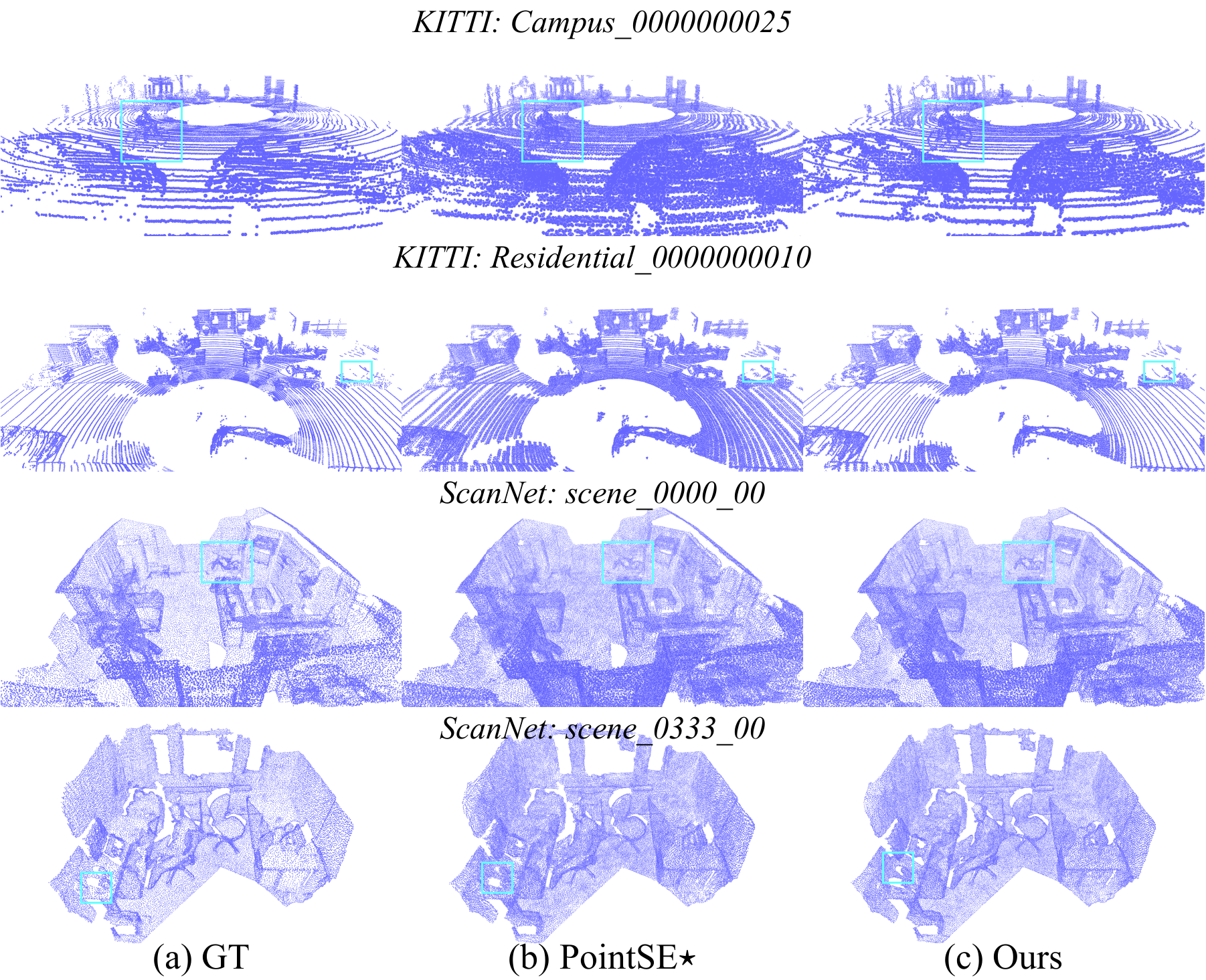}
  \caption{Recovery comparison on large scenes.}
  \label{fig:scene}
\end{figure}

\begin{table}[t]
\begin{center}
\scalebox{.7}{
\begin{tabular}{lcccccc}
\hline
 & \multicolumn{3}{c}{KITTI-10} & \multicolumn{3}{c}{ScanNet-30}  \\
\multirow{-2}{*}{Method} & CD & HD & EMD & CD & HD & EMD \\
\hline
PointSE$\star$ & 1.39 & 0.81 & 1.48 & 1.65 & 0.65 & 1.20 \\
Ours & \textbf{0.73} & \textbf{0.61} & \textbf{1.45} & \textbf{1.26} & \textbf{0.58} & \textbf{1.20}\\
\hline
\end{tabular}
}
\end{center}
\caption{CD ($10^5$), HD ($10^4$), and EMD ($10^4$) comparison on randomly selected scenes from KITTI and ScanNet.}\label{tab:scene}
\end{table}

\noindent \textbf{Scene-Level. }
On large scenes, we test the model trained on the random set using the collected scenes from KITTI-10 and ScanNet-30. We first normalize a scene and then divide it into patches with 2048 points. After recovery, the output patches are merged back together using their respective normalization parameters. Quantitative and qualitative results are demonstrated in Table~\ref{tab:scene} and Fig.~\ref{fig:scene}, where our method consistently outperforms PointSE$\star$, particularly with respect to outline appearance.

\begin{table}[t]
\begin{center}
\scalebox{.7}{
\begin{tabular}{lcccc}
\hline
Method & 4096 & 2048 & 1024 & 512 \\
\hline
SampleNet & 1.23/2.19 & 1.32/3.71 & 2.46/5.19 & 3.88/8.76 \\
FPS & 0.39/0.80 & 0.71/1.64 & 1.34/3.25 & 2.63/6.42 \\
APES & 0.45/1.22 & 0.85/2.41 & 1.64/4.69 & 3.25/9.17 \\
Ours & \textbf{0.36}/\textbf{0.75} & \textbf{0.69}/\textbf{1.51} & \textbf{1.31}/\textbf{2.99} & \textbf{2.54}/\textbf{5.88} \\
\hline
\end{tabular}
}
\end{center}
\caption{CD ($10^3$) errors of sampling methods at different resolutions using recovered/sampled points against GT.}\label{tab:sample}
\end{table}

\subsection{Point Cloud Sampling}\label{sec:sampling}

\begin{figure}[t]
  \centering
  \includegraphics[width=1.0\linewidth]{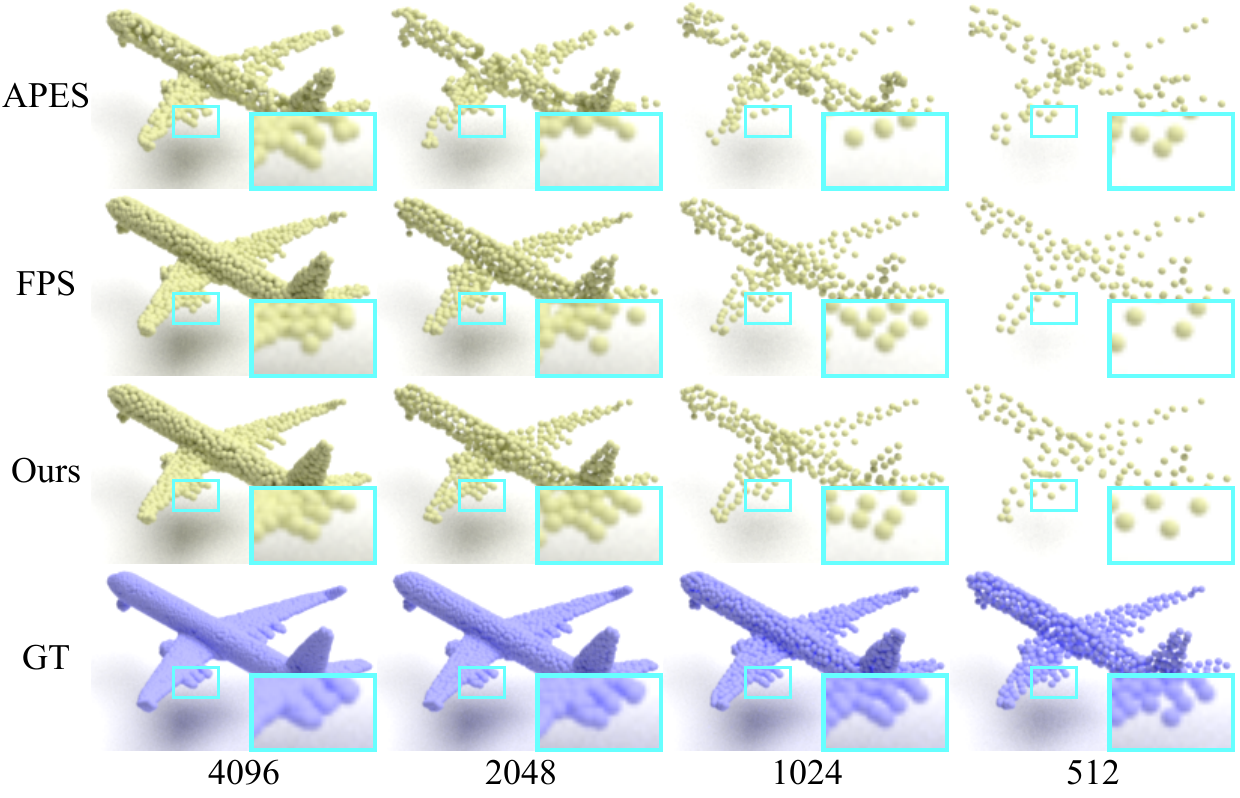}
  \caption{Sampling precisions. Track fairings are preserved.}
  \label{fig:sample}
\end{figure}

To demonstrate the superiority of our DP sampling module, we test on the random set of ModelNet40 at various resolutions: 4096, 2048, 1024, and 512, with a sampling ratio set to 4. For fair comparison, we use the same decoder and loss (i.e., without our $\mathcal{L}_\mathit{geo}$ and $\mathcal{L}_\mathit{top}$) of SnowflakeNet for all sampling methods. Moreover, sampling-only networks (i.e., SampleNet, our DP) are trained independently with GT as target if differentiable, and otherwise we directly compute CD of sampled points and GT from recovery networks (i.e., FPS, APES). In APES, we adjust the sampling ratios of its cascaded encoding layers be to 1 and 4 (ours 4 and 16), as we find that the original point number in the first layer yields the best results. As shown in Table~\ref{tab:sample}, among all sampling methods, our sampling significantly outperforms others across all resolution levels in both recovery and individual sampling tasks by 7.69\%/6.25\%, 2.82\%/7.93\%, 2.23\%/8\%, 3.42\%/8.41\%, respectively. This observation confirms not only the advantages of our designs to preserve topological and geometric properties, such as point-wise, point-shape and intra-shape correspondences, but also the significance of the sampling process in recovery.
Furthermore, we visualize the sampled results at different resolutions of GT in Fig.~\ref{fig:sample}, where the downsampling ratio is 4. Compared to APES and FPS, our sampling method preserves the clear track fairings on both wings of the aircraft, validating the preservation capacity of our method.

\begin{figure}[t]
  \centering
  \includegraphics[width=.98\linewidth]{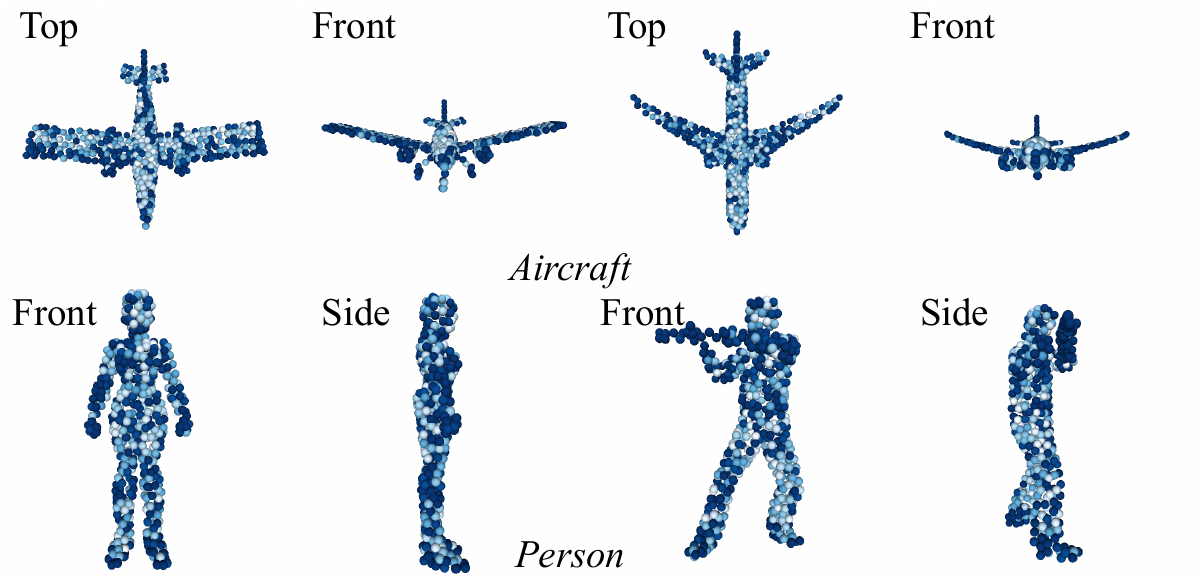}
  \caption{Predicted displacements. Darker blue represents larger offset. Regions with more details are generally more concentrated.}
  \label{fig:disp}
\end{figure}
\noindent \textbf{Displacement Prediction. }
Our network aims to learn displacements to preserve fine details, and thus we visualize the magnitude of those point movements in Fig.~\ref{fig:disp}. Different objects (aircraft and person) along with various instances are demonstrated to highlight consistency and symmetry in the results. For example, engines, wheels, and thin structures like wings in aircraft, as well as hands, feet, and tools in person are mostly concentrated by the method. These regions typically have smaller scales and higher geometric variance, which further demonstrates the preservation capacity of our sampling method.
\begin{table}[t]
\begin{center}
\scalebox{.7}{
\begin{tabular}{lccccccc}
\hline
Method & Param. & Train & Test & FLOPS & CD & HD & EMD \\
\hline
RepKPU & 1.02M & - & 19s(11s) & \underline{9}G & 0.99 & 0.35 & 3.07  \\
PointSE$\star$ & \underline{0.94M} & 4h & 30s(25s) & 41G & 0.86 & 0.34 & 2.97 \\
APES & 2.27M & 5h & 35s(29s) & 48G & 0.85 & 0.95 & 3.33 \\
SeedFormer & 2.27M & - & -  & - & - & - & - \\
Ours (S) & \textbf{0.46M} & 1.6h & 20s(12s) & \textbf{4}G & \underline{0.66} & \underline{0.29} & \textbf{2.78} \\
Ours & 2.03M & 3h & 24s(14s) & 14G & \textbf{0.62} & \textbf{0.29} & \underline{2.82} \\\hline
\end{tabular}
}
\end{center}
\caption{Architecture complexity comparison. }\label{tab:complexity}
\end{table}

\subsection{Complexity Analysis}

To demonstrate the efficiency of our method, Table~\ref{tab:complexity} compares architecture parameters and FLOPS (G). Testing times include evaluations for individual metrics (CD, HD, and EMD). Notably, SeedFormer is only referenced for model size comparison, as it requires additional seed features during decoding. Evaluation times are indicated in parentheses. PointSE$\star$ has more feature extraction modules, consisting of multiple FPS sampling and kNN grouping operations. While they do not contain any parameters, they extremely increase the computational cost. Thus, our training time is 25\% faster than PointSE$\star$. We also include our small (S) model to show that performance is still better than PointSE$\star$ with fewer parameters. In summary, by eliminating redundant grouping calculations and leveraging the proposed ITA module (only 0.08M parameters), our model achieves the highest efficiency in both training and testing phases.

\subsection{Ablation Study}\label{sec:ab}
\begin{table}[t]
\begin{center}
\scalebox{.7}{
\begin{tabular}{ccc|cc|cc|ccc}
\hline
\multicolumn{3}{c}{Sampling} & \multicolumn{2}{c}{Restoration} & \multicolumn{2}{c}{Loss} & \multicolumn{3}{c}{Metric}\\
 $P_d$ & $\delta_d$ & ITA & SPD & UP & $\mathcal{L}_\mathit{geo}$ & $\mathcal{L}_\mathit{top}$ & CD & HD & EMD \\
\hline
 $\surd$ & - & $\surd$ & $\surd$ & - & $\surd$ & - & 0.68 & 3.00 & \textbf{2.68} \\
 $\surd$ & $\surd$ & - & $\surd$ & - & $\surd$ & - & 0.65 & 2.90 & 2.75 \\
 $\surd$ & $\surd$ & $\surd$ & $\surd$ & - & - & - & 0.69 & 3.06 & 2.81 \\
 $\surd$ & $\surd$ & $\surd$ & $\surd$ & - & $\surd$ & - & 0.64 & 2.91 & 2.77 \\
 $\surd$ & $\surd$ & $\surd$ & $\surd$ & - & $\surd$ & $\surd$ & 0.63 & 2.91 & 2.83 \\
 $\surd$ & $\surd$ & $\surd$ & - & $\surd$ & $\surd$ & - & 0.64 & 2.89 & 2.75 \\
 $\surd$ & $\surd$ & $\surd$ & - & $\surd$ & $\surd$ & $\surd$ & \textbf{0.62} & \textbf{2.89} & 2.82 \\
\hline
\end{tabular}
}
\end{center}
\caption{Ablation study on the random set of ModelNet40 in terms of CD ($10^3$), HD ($10^3$), and EMD ($10^2$).}\label{tab:ab}
\end{table}

We elaborate the ablation study in Table~\ref{tab:ab}, where we evaluate the contributions of various components in both the sampling and recovery phases. In the sampling phase, we highlight the importance of the geometric and topological embeddings, as well as the proposed ITA. In the restoration phase, we compare the performance of different decoding modules, including SPD~\cite{xiang2021snowflakenet} and our proposed UP. 
Note that, neither seed features~\cite{zhou2022seedformer} nor global code~\cite{xiang2021snowflakenet} is transmitted to our UP-based restoration phase. At last, we assess the significance of the proposed losses, $\mathcal{L}_\mathit{geo}$ and $\mathcal{L}_\mathit{top}$. Specifically, the result of 2048 points without our proposed losses is also provided in Table~\ref{tab:sample}. With these designs, our overall architecture achieves the lowest CD and HD but slightly higher EMD. This is because our predicted displacement prioritizes preserving shape and boundaries over enforcing uniformity, which is empirically a crucial factor for EMD.

\section{Conclusion and Discussion}

Unlike existing sampling and upsampling methods that focus on individual tasks, we propose an end-to-end network, \textit{TopGeoFormer}, for point cloud recovery. By encoding both topological and geometric attributes, our method achieves outstanding performance in both recovery and sampling tasks. Extensive experiments demonstrate that our method deliberately rearranges points for precise shape coverage, leading the state-of-the-art. Our model can be deployed in remote systems for energy-efficient processing and environmental monitoring. We expect this study will serve as a foundational contribution to future recovery research. 

\noindent \textbf{Limitation}: \textit{TopGeoFormer} is designed to favor non-uniform inputs and assumes none or low levels of noise. Moreover, since it is trained solely on complete objects, it is not suited for predicting largely missing regions.

\section*{Acknowledgments}
This work is supported by the research fund under Grant No. 20242910035 from the Tsinghua University-Jiangsu CRRC Digital Technology Co.,Ltd. Joint Research Center for Data Driven Intelligence of Industry.


\end{document}